\pdfoutput=1

\documentclass[11pt]{article}

\usepackage{acl}

\usepackage{times}
\usepackage{latexsym}

\usepackage[T1]{fontenc}

\usepackage[utf8]{inputenc}

\usepackage{microtype}
\usepackage{times}
\usepackage{soul}
\usepackage{url}
\usepackage{hyperref}
\usepackage[utf8]{inputenc}
\usepackage{caption}
\usepackage{graphicx}
\usepackage{subfigure}
\usepackage{amsmath}
\usepackage{amsthm}
\usepackage{booktabs}
\usepackage{algorithm}
\usepackage{algorithmic}
\usepackage{multirow}
\usepackage{footnotehyper}
\usepackage{color}
\usepackage{CJKutf8}
\title{Pretraining without Wordpieces: \\Learning Over a Vocabulary of Millions of  Words}

\author{
Zhangyin Feng, Duyu Tang, Cong Zhou, Junwei Liao, Shuangzhi Wu,\\
\textbf{Xiaocheng Feng, Bing Qin, Yunbo Cao, Shuming Shi }\\
}

\begin{document}
\maketitle
\begin{abstract}
The standard BERT adopts subword-based tokenization, which may break a word into two or more wordpieces (e.g., converting ``\textit{lossless}'' to ``\textit{loss}'' and ``\textit{less}''). This will bring inconvenience in following situations: (1) what is the best way to obtain the contextual vector of a word that is divided into multiple wordpieces? (2)  how to predict a word via cloze test without knowing the number of wordpieces in advance? 
In this work, we explore the possibility of developing BERT-style pretrained model over a vocabulary of words instead of wordpieces.
We call such word-level BERT model as WordBERT.
We train models with different vocabulary sizes, initialization configurations and languages.
Results show that, compared to standard wordpiece-based BERT, WordBERT makes significant improvements on cloze test and machine reading comprehension.
On many other natural language understanding tasks, including POS tagging, chunking and NER, WordBERT consistently performs better than BERT.
Model analysis indicates that the major advantage of WordBERT over BERT lies in the understanding for low-frequency words and rare words.
Furthermore, since the pipeline is language-independent, we train WordBERT for Chinese language and obtain significant gains on five natural language understanding datasets.
Lastly, the analyse on inference speed illustrates WordBERT has comparable time cost to BERT in natural language understanding tasks.
\end{abstract}

\section{Introduction}
\begin{figure*}[ht]
\begin{center}
\includegraphics[width=\linewidth]{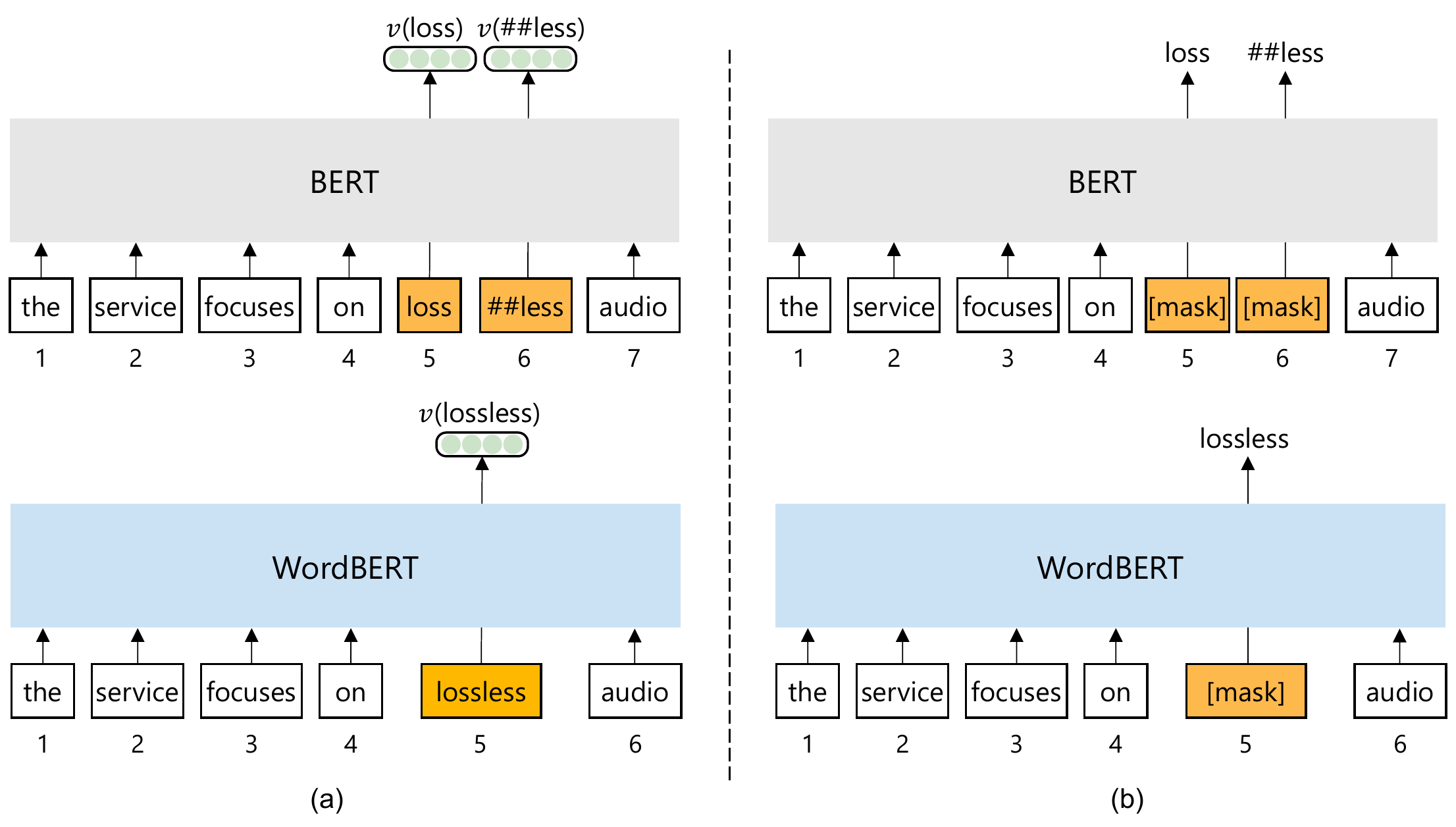}
\caption{(a) BERT only produces representations for subwords \textit{``loss"} and \textit{``\#\#less"}, while WordBERT is able to directly obtain the representation for \textit{``lossless"}. 
(b) BERT is unable to generate a word that divided into multiple subwords (e.g., \textit{``lossless"}) on top of one \texttt{[mask]} token.
In order to generate the word \textit{``lossless"} via cloze test, BERT needs to know the oracle number of subwords. In contrast, WordBERT inherently escapes this dilemma to predict a word based on one \texttt{[mask]} token.}
\label{figure-question}
\end{center}
\end{figure*}

Representations of words work as the foundation of modern natural language processing systems \cite{bommasani2021opportunities}. 
The journey of pretrained word representations starts from context-independent word embeddings \cite{bengio2003neural,mikolov2013efficient, pennington2014glove}, where the representation of a word remains the same in different contexts.
Different from word embeddings, pretrained models like BERT \cite{bert} and RoBERTa \cite{liu2019roberta} produce representations of words conditioned on contexts.
They are typically built with Transformer \cite{vaswani2017attention} and 
use sub-word based tokenizers (e.g., WordPiece \cite{wu2016google}) that decompose a word into small subword units. For example, a word ``\textit{lossless}'' may be divided into two subwords ``\textit{loss}'' and ``\textit{less}'' before consumed into the Transformer.
Although pretrained models achieve awesome performance on a wide range of downstream tasks, the use of subwords brings inconvenience in following scenarios.
The first scenario is to compute the contextual representation of a word. For example, in Figure ~\ref{figure-question} (a), the word \textit{``lossless"} is divided into two tokens. Although a commonly approach is to use the vector of the first subword (i.e., \textit{``loss"}) as the vector of the whole word, it is suboptimal to ignore the contextual representations of remaining subwords.
The second scenario is to predict a word given surrounding texts (i.e., cloze test).
As illustrated in Figure ~\ref{figure-question} (b), given the context of ``\textit{the service focuses on \underline{\ \ \ \ \  } audio}'', the model struggles to predict a compound word (e.g., ``\textit{lossless}'') before knowing the number of wordpieces of the target word. 

In this work, we explore the learning of BERT-style pretrained model that discards subword-based tokenization, so that the aforementioned issues are essentially avoided. 
The goal of this work is not to propose a new neural network architecture or a new pretraining objective. Instead, we aim to explore whether it is possible to build an alternative BERT-style pretrained that has the merits of learning contextual representations of words from vast unlabeled data and retaining the integrality of words. 
We largely follow the model architecture of BERT and develop on top of Transformer. 
We call this word-level pretrained model as WordBERT.
Since the whole pipeline is language-ignostic, we train WordBERT for both English and Chinese languages.
For English language, we also explore various configurations, such as different vocabulary sizes (e.g., 500K and 1M) and whether considering off-the-shell word embeddings like Glove.
To alleviate the huge computation of softmax for large vocabulary, we adopt a sampling-based training strategy.
During training, instead of updating all word embedding parameters, we dynamically sample a small part of word embedding parameters to update in each batch.

To comprehensively evaluate the performance of WordBERT, we conduct experiments on many tasks, including cloze test, machine reading comprehension, Part-Of-Speech (POS) tagging, chunking and named entity recognition (NER).
Experimental results indicate that WordBERT gains a huge performance boost over the BERT on cloze test and machine reading comprehension, and
 consistently outperforms BERT on POS tagging, chunking and NER tasks.
To further investigate which types of words benefit from our model, we construct a probing analysis to test WordBERT in a zero-shot scenario without fine-tuning the parameters. 
The analysis results illustrate that WordBERT has a better understanding for low-frequency words and rare words than BERT.
Furthermore, we also train a model for Chinese to verify the effectiveness of our approach on other languages, and find that our model brings significant improvements on the standard CLUE benchmark \cite{xu2020clue} for Chinese.
More importantly, we find that although WordBERT has larger model parameters than BERT, it has comparable time cost to BERT when being applied to 
natural language understanding tasks.

\begin{table*}[t]
\centering
\setlength{\tabcolsep}{3pt}
\begin{tabular}{l|c|c|c|c|c}
\toprule
\bf Configurations & BERT & WordBERT-500K  & WordBERT-1M & WordBERT-Glove & WordBERT-ZH \\
\midrule
Vocab Size & 30.5K &500K & 1M & 1.9M & 278K \\
Language & English &English& English & English & Chinese \\
Transformer Params& 85M & 85M  & 85M & 85M & 85M  \\
Embedding Params & 23.4M & 384M  & 768M  & 570M & 213M \\
Embedding Dim & 768 & 768 & 768 & 300 & 768\\
\bottomrule
\end{tabular}
\caption{Models with different configurations.}
\label{tab:models}
\end{table*}

\section{Background}
In this section, we give a brief overview of subword-based tokenization and BERT ~\cite{bert} pretraining approach.

The default tokenization used in recent pre-training models \cite{bert,lewis-etal-2020-bart} is subword-based tokenization.
The subword-based tokenization algorithms uses the following principles: (1) Do not split the frequently used words into smaller subwords. (2) Split the low-frequency words into smaller meaningful subwords.
Wordpiece \cite{wu2016google}, used in BERT, is one of the subword tokenizations.
As shown in Figure~\ref{figure-question}, given a text sequence \textit{``the service focuses on lossless audio"}, wordpiece model will generate the corresponding wordpiece sequence \textit{``the service focuses on loss \#\#less audio"}.

BERT contains two components: embedding layer and Transformer.
Given an input, BERT first gets the corresponding static embedding representations through embedding layer, and then produces the contextual representations through stacked Transformer encoders. 
During pretraining, BERT uses two objectives: masked language modeling (MLM) and next sentence prediction (NSP).
The MLM objective is a cross-entropy loss on predicting the masked tokens.
A random sample of the tokens in the input sequence is selected and replaced with the special token {\tt [MASK]}.
NSP is a binary classification loss for predicting whether two segments follow each other in the original text.
\section{Model}

In this section, we introduce our models and the sampling-based training processing.
As shown in Table~\ref{tab:models}, we train several models with different configurations. 
We'll go into the details next.

\subsection{WordBERT}
Similar with BERT, WordBERT contains two components: word embedding and Transformer layer.
Word embedding is a parameter matrix to obtain the vector representations of words.
Different with BERT \cite{bert} using subword units, WordBERT’s vocabulary is made up of complete words without wordpieces.
In this work, we use spacy\footnote{\href{https://spacy.io/}{https://spacy.io/}} to process the data, and take the top K words with the highest frequency as the vocabulary.
Specifically, we train two models with vocabulary sizes of 500K and 1M, named WordBERT-500K and WordBERT-1M respectively.
In addition, we add five special words (\textsl{[PAD], [UNK], [CLS], [SEP] and [MASK]}) to the vocabulary.
Following previous models \cite{bert,liu2019roberta}, we adopt the multi-layer bidirectional transformer \cite{vaswani2017attention} to learn contextualized representation.
The transformer architecture has $L$ layers. Each block uses $A$ self-attention heads and hidden dimension $H$.

\subsection{WordBERT-Glove}
The parameters of word embedding in WordBERT-500K and WordBERT-1M are randomly initialized.
To explore the effectiveness of using pretrained word vectors, we adopt the Glove vocabulary \cite{pennington2014glove} with approximately 1.9 million uncased words as the vocabulary and use the corresponding word vectors to initialize our model, named WordBERT-Glove.
Glove embedding is pre-trained on Common Crawl dataset with 42B tokens \footnote{\href{https://nlp.stanford.edu/projects/glove/}{https://nlp.stanford.edu/projects/glove/}} with 300 dimension.
Given a sequence of words, we first get the corresponding 300-dimensional representation for each word through Glove Embedding.
Then, we add a linear layer to map the 300-dimensional feature space of Glove to the 768-dimensional feature space of the transformer layer.

In order to better map the feature from glove embedding space to transformer space, we pretrain the linear layer for WordBERT-Glove.
Specially, we extract the identical words in the Glove and BERT vocabulary and the corresponding 300-dimensional and 768-dimensional vectors. 
There are 22, 860 overlapping words in total.
We set the 300-dimensional Glove vector as input  $v_{in}$ and set the 768-dimensional BERT vector as output $v_{out}$.
The linear layer is trained  with the mean squared error loss function as the follows:
\begin{align}
\hat{v}_{out} & = W^{T}*v_{in} \\
\mathcal{L}_{mse} & = \frac{1}{N} \sum_{i=1}^{N} (\hat{v}_{out} - v_{out})^{2}
\end{align}
where $W$ is the parameters of the linear layer and N is the number of overlapping words.
\subsection{WordBERT-ZH}
Considering that our method is language-independent, we also train a Chinese version model, named WordBERT-ZH, to verify the effectiveness of word-based vocabulary in Chinese language.
Similar with WordBERT-500K and WordBERT-1M, WordBERT-ZH's vocabulary is obtained from the training corpus.
We condider WuDaoCorpus~\cite{YUAN202165} \footnote{\href{https://resource.wudaoai.cn/home}{https://resource.wudaoai.cn/home}}, a 200GB raw text from various domains.
Furthermore, we use spacy to process the corpus and build the vocabulary with the size of 278K.
The parameters of word embedding are randomly initialized and we keep the dimension of word embedding as 768. 
\begin{savenotes}
\begin{table}[t]
\centering
\begin{tabular}{lccc}
\toprule
\bf Model  & CLOTH & M  &  H \\
\midrule
LSTM$^ \dagger $ & 48.4 & 	51.8	& 47.1 \\
BERT-base$^\dagger$ & 82.0	 & 85.0 &	80.9  \\
Human$^\dagger$  & 85.9 & 89.7 & 	84.5 \\
\midrule
WordBERT-500K & 84.99 & 	87.40 &	84.06 \\
WordBERT-1M & \bf85.06 &	\bf87.59 &	\bf84.08  \\
WordBERT-Glove & 83.26 &	86.27 &	82.10 \\
\bottomrule
\end{tabular}
\caption{Evaluation results on the CLOTH dataset. 
M and H are shorthand for CLOTH-M and CLOTH-H, respectively.
$\dagger$ indicates the results are from leaderboard. \footnote{\url{http://www.qizhexie.com/data/CLOTH\_leaderboard.html}}}
\label{tab:cloze-results}
\end{table}
\end{savenotes}

\subsection{Model Training}
Following BERT-base \cite{bert}, we set the number of Transformer layers as 12, the hidden size as 768, the number of self-attention heads as 12 and the masked ratio as 15\%.
Different with the original BERT, which is trained with masked language modeling (MLM) task and next sentence prediction (NSP) task, we only employ MLM task to train our model and remove NSP task, which is proved ineffective in RoBERTa \cite{liu2019roberta}.
The pre-training data are BookCorpus \cite{zhu2015aligning} and English Wikipedia, which are the original data used to train BERT.
For different versions of our models, the parameters of the transformer part are initialized with BERT.
We list the differences between different configuration models in Table~\ref{tab:models}.
For WordBERT-500K, WordBERT-1M and WordBERT-ZH, the embedding dimension is 768 and the parameters of embedding are randomly initialized and .
For WordBERT-Glove, the parameters of embedding are initialized with Glove embedding and the dimension of word embedding is 300. 
We initialize the linear layer with the pretrained parameters and fix the word embedding parameters unchanged.

To reduce the huge computation cost of softmax and speed up training, we dynamically sample a small vocabulary for each batch during pretraining.
Specifically, for each batch, our new vocabulary contains two parts, including (1) randomly sample 30,000 words from the entire vocabulary; (2) all words contained in the current batch. For WordBERT-Glove, the sampled small vocabulary additionally includes the 10 most similar words for each masked word in the vocabulary, which are obtained through Glove embedding similarity by an off-the-shelf toolkit Annoy.\footnote{\href{https://github.com/spotify/annoy}{https://github.com/spotify/annoy}}
Detailed hyper-parameters for model training are given in appendix~\ref{appendix-pretraining}.

\section{Experiment}
We present empirical results in this section to verify the effectiveness of our models.
We evaluate our models in various tasks, such as cloze test, machine reading comprehension, POS tagging, Chunking and NER.
After that, we construct two probing datasets to analyze the model’s performance on different frequencies words. 
Furthermore, we evaluate our Chinese model WordBERT-ZH on CLUE benchmark \cite{xu2020clue}.
Lastly, we compare the inference speed on different tasks and give several cases for probing to analyze our models.

\begin{table*}[t]
\centering
\begin{tabular}{l|cc|cc|cc|cc}
\toprule
\multirow{2}{*}{\bf Model}  & \multicolumn{2}{c|}{POS Tagging} & \multicolumn{2}{c|}{Chunking}  &  \multicolumn{2}{c|}{CoNLL2003} & \multicolumn{2}{c}{CoNLL++}  \\
  & Dev &Test & Dev & Test & Dev &Test & Dev & Test  \\
\midrule
BERT-base  & 93.12 & 91.96 & 92.38 & 91.18 & 95.18 & 90.92 & 95.18 &91.96   \\
WordBERT-500K  & 93.11 & 91.92 & 92.44  &91.36 & 95.24  & 91.54  & 95.24 & 92.43   \\
WordBERT-1M  & 93.08 & 92.02 & \bf 92.50  & 91.36   &\bf95.39  & \bf91.66 & \bf95.39 & \bf92.70 \\
WordBERT-Glove & \bf93.77 & \bf93.32& 92.49 & \bf91.52 & 95.27 & \bf91.66& 95.27& 92.65\\
\bottomrule
\end{tabular}
\caption{Evaluation results on the CoNLL2003 dataset for POS tagging and Chunking.}
\label{tab:chunk-pos-results}
\end{table*}

\subsection{Cloze Test}
Cloze test \cite{taylor1953cloze} is the task of infilling in blank positions of text which are consistent with preceding and subsequent text, which is a classic language exercise in English exams.
We adopt CLOTH \cite{xie-etal-2018-large}, a large-scale cloze test data set created by teachers, to evaluate our models.
Questions in the data set are designed by middle-school and high school teachers to prepare Chinese students for entrance exams.
CLOTH is divided into CLOTH-M and CLOTH-H, which stand for the middle school part and the high school part.
This data set has four types of questions, including 
(1) Grammar: the question is about grammar usage involving tense, preposition usage, subjunctive mood, active/passive voices and so on; 
(2) Short-term-reasoning: the question can be answered based on the information within the same sentence; 
(3) Matching: the question is answered by paraphrasing a word in the context; 
(4) Long-term-reasoning: the answer must be inferred from synthesizing information distributed across multiple sentences. 
With missing blanks and candidate options carefully created by teachers to test different aspects of language phenomena, CLOTH requires a deep language understanding and better captures the complexity of human language.

Follow the experimental settings of the original paper \cite{xie-etal-2018-large}, we formulate the problem of cloze test as a multi-choice question answering task, where the question is a passage in which some words are replaced by blanks and distractor candidate answers are incorrect but nuanced to make the test non-trivial. 
Detailed hyper-parameters for model fine-tuning are given in appendix~\ref{appendix-finetuning}.

Results are given in Table \ref{tab:cloze-results}. We report accuracy, namely the number of correctly predicted questions over the number of all questions, for each dataset.
The results of LSTM, BERT-base and Human are from leaderboard.
Overall, various versions of our model performer better than LSTM and  BERT-base.
Specifically, WordBERT-500K, WordBERT-1M and WordBERT-Glove achieve accuracy scores of 84.99, 85.06, and 83.26, which exceeds BERT-base 2.99 and 3.06, 1.26 points respectively.
What's more, WordBERT-1M improves by 3.19 points over BERT-base on the high school dataset than that 2.59 points on the middle school dataset, which indicates our models have better understanding and reasoning ability, especially for complex questions.

\subsection{Sequence Labeling}
We also evaluate our models on several sequence labeling tasks, such as Part-Of-Speech (POS) tagging, Chunking and named entity recognition (NER).
All these tasks can be regarded as word-level classification tasks. A common practice for sub-word based BERT is to use the hidden state of the first sub-token to represent the whole word, which may hurt the model performance.
We use CoNLL2003 \cite{sang2003introduction} for POS tagging and Chunking with standard split. For NER task, in addition to CoNLL2003, we also use CoNLL++ \cite{wang2019cross} which the mistakes in the test set are manually corrected.
We report $F_{1}$ on development and test datasets. 

Results are shown in Table~\ref{tab:chunk-pos-results}. 
We find that our models can achieve slightly better results on POS tagging and Chunking tasks.
Compared with POS tagging and Chunking, our models are able to achieve more significant improvements on NER task.
We speculate that our models may have advantage in learning representations for low-frequency words since named entities tend to be infrequent words.

\begin{table}[t]
\centering
\setlength{\tabcolsep}{3pt}
\begin{tabular}{l|cc|cc}
\toprule
\multirow{2}{*}{\bf Model} &\multicolumn{2}{c|}{SQuAD 1.1} & \multicolumn{2}{c}{SQuAD 2.0} \\
  &  EM & $F_{1}$ &   EM  &$F_{1}$ \\
\midrule
BERT-base & 80.8 &	88.5 &	73.3 &	76.3 \\
WordBERT-500K & 82.31 & 89.62 & 73.32 & 76.22 \\
WordBERT-1M & 82.42 &	89.57 & 73.73 & 76.90  \\
WordBERT-Glove & \bf82.69 & \bf89.78 & \bf74.06 & \bf77.18  \\
\bottomrule
\end{tabular}
\caption{Evaluation results on the SQuAD development datasets.}
\label{tab:squad-results}
\end{table}

\begin{table*}[]
    \centering   
    \begin{tabular}{l|cccc|cccc}
        \toprule
        \multirow{2}{*}{\bf Model} & \multicolumn{4}{c|}{SQuAD} & \multicolumn{4}{c}{WikiText-103} \\
         & High & Medium & Low & Rare & High & Medium & Low & Rare   \\
        \midrule
        BERT-base & 59.31 & 25.21 & 6.20  & 0.00 & 57.28 &21.20 & 3.28  & 0.00 \\
        WordBERT-500K & 61.80 & \bf35.68 & 27.28 & 4.78 & 59.87 &\bf35.67 & 	23.30 &	3.31 \\
        WordBERT-1M & 61.31 &	35.56 &	28.25 &	\bf4.95 &	59.41 &	34.09 & 24.09 & \bf4.19 \\
        WordBERT-Glove & \bf63.23 & 34.95 & \bf28.54& 4.30& \bf61.30 &33.99& \bf25.20	&3.98 \\
        \bottomrule
    \end{tabular}
    \caption{Evaluation results on probing test for different frequencies  words. Accuracies (\%) are reported. Best results in each group are in bold.}
    \label{tab:wordinfilling-results}
\end{table*}

\subsection{SQuAD}
The Stanford Question Answering Dataset (SQuAD) provides a paragraph of context and a question. The task is to answer the question by extracting the relevant span from the context. We evaluate on two versions of SQuAD: v1.1 \cite{rajpurkar2016squad} and v2.0 \cite{rajpurkar-etal-2018-know}. In SQuAD v1.1 the context always contains an answer, whereas in SQuAD v2.0 some questions are not answered in the provided context, making the task more challenging.
Following BERT, we represent the input question and passage as a single packed sequence and fine-tune the model to find the start word index and the end word index of the answer span.
For SQuAD v2.0, we treat questions that do not have an answer as having an answer span with start and end at the \textsl{[CLS]} token.
Detailed hyper-parameters for model fine-tuning are given in Appendix~\ref{appendix-finetuning}.

Table~\ref{tab:squad-results} shows the performance of various models, where exact match (EM) ratio and $F_{1}$ are reported  on development dataset. 
According to the table, we can see that our models are superior in both SQuAD v1.1 and SQuAD v2.0 overall. 
Specifically, WordBERT-Glove achieves 89.69 EM score and 89.78 $F_{1}$ score,
which illustrates that our models have better understanding of the context and question, and accurately identify the correct spans.

\subsection{Probing}
To deeply explore the relation between the model performance and the frequency of a word during pretraining, we split test set into four subsets based on the word frequency in WikiText-103. We set 3000, 300, and 3 as the thresholds to divide words into high-frequency words (High), medium-frequency words (Medium), low-frequency words (Low) and rare words (Rare), respectively.
We build two probing datasets with different sizes, which are constructed based on WikiText-103 and SQuAD v1.1.
We select 200K sentences from WikiText-103 and all 19K contexts from SQuAD v1.1.
We separately mask out different frequencies words with a random 15\% probability, and then let the model predicts the word in a zero-shot setting where the parameters are fixed.
Statistics of the data for probing is given in \ref{appendix-finetuning}.

Results are given in Table~\ref{tab:wordinfilling-results}. 
We report top 1 accuracy for two probing datasets across different frequencies subsets.
Results show that our models outperform than BERT across all frequency words on both WikiText-103 and SQuAD probing dataset.
As can be seen, the performance of all models decrease significantly with the word frequency decreasing.
For high frequency words, our models bring limited improvements, indicating that both BERT and our models can learn good representations for high-frequency words.
Besides, our models significantly outperform BERT for other three types of words. For instance, WordBERT-500K and BERT-base achieve 35.68 and 25.21 top 1 accuracy on the mid-frequency subset of SQuAD, demonstrating that our models are able to learn better representation for medium-frequency words, low-frequency words and rare words.

Although WordBERT suffers from out-of-vocabulary (OOV) problem,
we believe if the vocabulary size is large enough, only a few least-known words are not included in the vocabulary.
In practice, due to the few occurrences in pretraining corpus, subword-based models may not learn good representations for those OOV words either. We show that our model has better learning ability for low-frequency words and rare words than BERT.

\begin{savenotes}
\begin{table*}[]
    \centering   
    \begin{tabular}{l|c|c|c|c|c}
    \toprule
     \bf Model & TNEWS & IFLYTEK & AFQMC & OCNLI & CSL  \\
        \midrule
        BERT-base  $^\ddagger$  & 56.09 & 60.37 & 74.16 & 74.70 &  79.63 \\
        RoBERTa-base  $^\ddagger$ & 57.51 & 60.80 & 74.30 & 75.01 & 80.67  \\
        WoBERT   $^ \dagger $ & 57.01 & 61.10 & 72.80 & 75.00 & - \\
        MarkBERT   $^ \dagger $ & 58.40 & 60.68 & \bf74.89 & 75.88 &  - \\
        \midrule
        WordBERT-ZH & \bf59.75 	& \bf61.56 &	74.64 &	\bf77.63 &	\bf83.33 \\
        \bottomrule

    \end{tabular}
    \caption{Evaluation results on Chinese NLU datasets in terms of accuracy (\%). 
    $^ \ddagger $ indicates the results are from leaderboard. \footnote{ \url{https://github.com/CLUEbenchmark/CLUE}}
    $^ \dagger $ indicates the results are from \citet{markbert}.
    }
    \label{tab:clue}
\end{table*}
\end{savenotes}

\subsection{Chinese NLU}
We use various types of language understanding tasks from the CLUE benchmark \cite{xu2020clue} to evaluate WordBERT-ZH.
Specifically, we use classification tasks such as TNEWS, IFLYTEK; semantic similarity task (AFQMC); keyword recognition (CSL); natural language inference task (OCNLI).
In addition to BERT \cite{bert} and RoBERTa \cite{cui2019pre}, we also draw WoBERT~\cite{zhuiyiwobert} and MarkBERT~\cite{markbert} as baselines.
WoBERT is a Chinese pre-trained model initialized from the BERT BASE pre-trained weights, which has a 60K expanded vocabulary containing commonly used Chinese words.
MarkBERT remains the vocabulary being Chinese characters and inserts boundary markers between contiguous words.
We follow the implementation details used in the CLUE benchmark official website and report the best accuracy on valid datasets.

As we can see in Table~\ref{tab:clue}, WordBERT-ZH outperforms baselines on five tasks and achieves significant improvements on TNEWS, OCNLI and CSL tasks.
Specifically, WordBERT-ZH improves by 3.66 points for TNEWS, 2.93 points for OCNLI and 3.7 points for CSL over BERT-base. 
This reflects that the word-based model is also very effective for Chinese language.

\begin{figure}[ht]
\centering
\includegraphics[width=\linewidth]{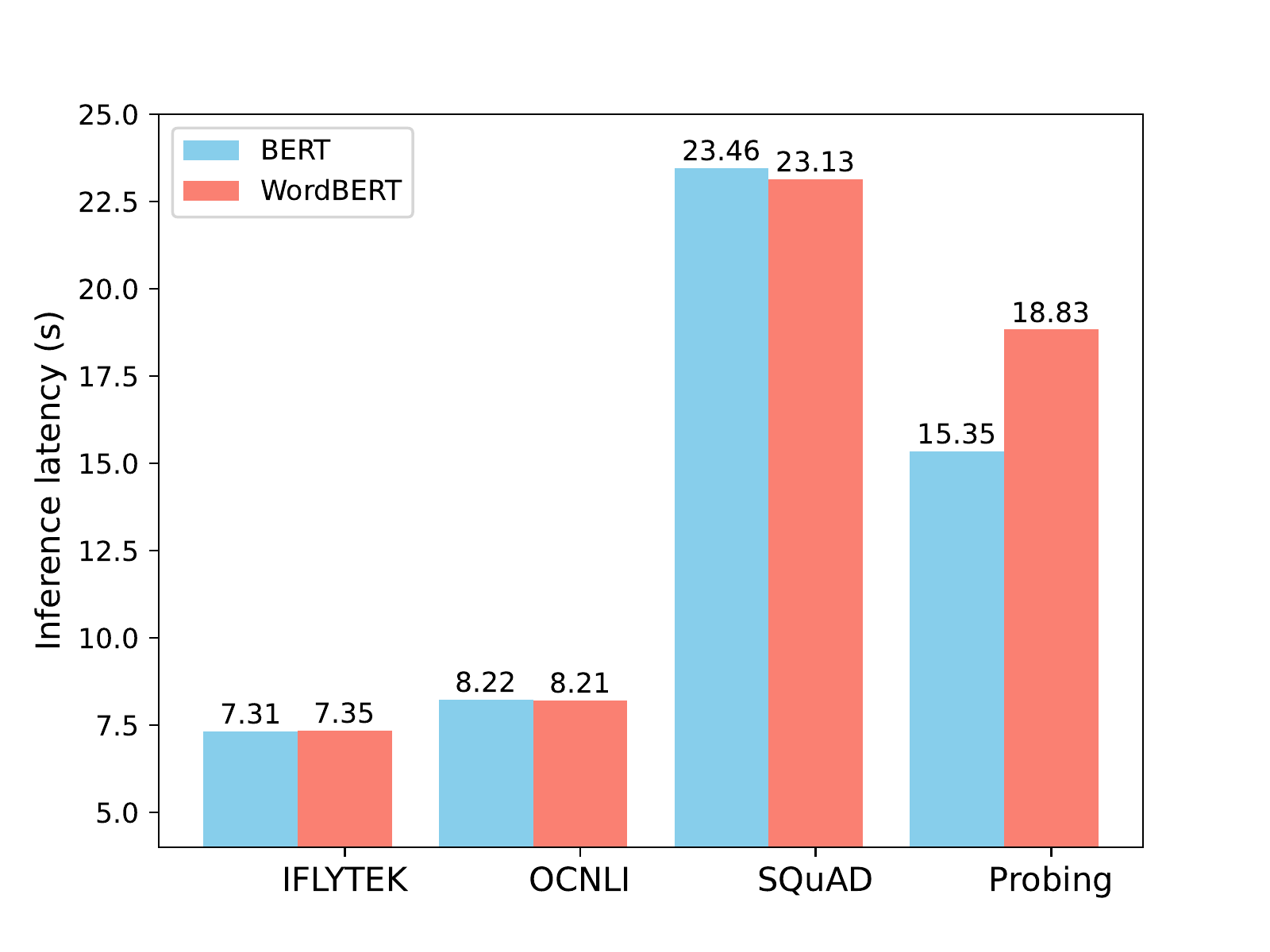}
\caption{Inference latencies on different tasks.}
\label{fig-speed}
\end{figure}

\begin{figure*}[ht]
\centering
\includegraphics[width=\linewidth]{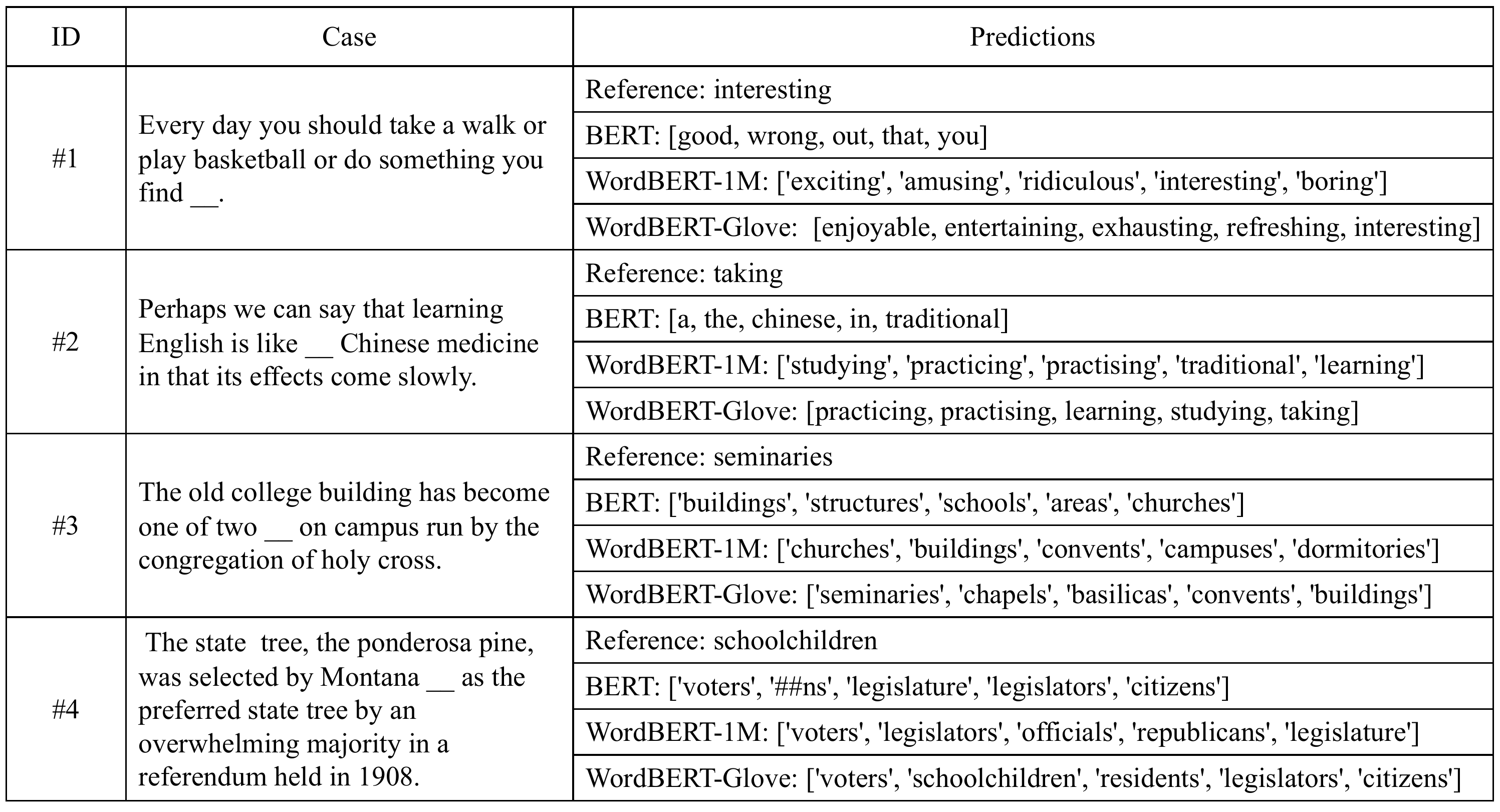}
\caption{Case study on probing. We give the reference word and the top-5 generated words for each case. The reference words of case \#1 and case \#2 are high-frequency words, and the reference words of case \#3 and case \#4 are low-frequency words.}
\label{fig-case}
\end{figure*}

\subsection{Inference Speed}
We compare the inference speed of BERT-base and WordBERT on different tasks.
All inference speed tests are performed on one same NVIDIA machine. 
For IFLYTEK, OCNLI and SQuAD tasks, we report the inference time (in seconds) of conducting model inference on the development set for one round.
For the probing task, we report the inference time on a subset of 1,000 contexts from SQuAD high frequency dataset. We use WordBERT-ZH for IFLYTEK and OCNLI tasks, WordBERT-500K for SQuAD and probing tasks.

As shown in Figure~\ref{fig-speed}, on NLU tasks (i.e., the first groups),
the overall inference time of WordBERT and BERT is comparable.
The reason is that, even though our vocabulary size is significantly larger,  obtaining word embedding is a dictionary lookup operation in python, which is independent of the size of vocabulary.\footnote{\href{https://wiki.python.org/moin/TimeComplexity}{https://wiki.python.org/moin/TimeComplexity}}
Moreover, we find WordBERT has even slightly faster inference speed for SQuAD and OCNLI tasks.
The reason might be that the length of the input sequence without sub-words is shorter, which reduces the computational cost of self-attention.
For the probing task, WordBERT takes more time than BERT due to the computational cost of softmax over a large vocabulary.

\subsection{Case Study}
In this section, we give several examples for probing to qualitatively analyze our models.
Given a sentence with a blank, we first replace the blank with \textsl{[MASK]}, and then use model to predict a word to fill in the blank.
We directly adopt pretrained BERT, WordBERT-1M and WordBERT-Glove without modifying the parameters.

As shown in Figure~\ref{fig-case}, for each case, we give the top-5 words generated by models and the reference word.
The case \#1 and case \#2 correspond to high-frequency words, and the case \#3 and case \#4 correspond to low-frequency words.
From case \#1 and case \#2, we can find that our models are able to generate more diverse and meaningful words, while BERT tends to produce general words.
In case \#3 and case \#4, our models are able to produce some uncommon words, while BERT generates frequent words or even subwords, such as \textit{``\#\#ns"} in case \#4.

\section{Related Work}
Traditional word vector approaches were built upon word-level vocabularies with millions of words, such as Word2vec~\cite{mikolov2013efficient}, Glove \cite{pennington2014glove} and ELMo \cite{peters-etal-2018-deep}.
However, Word2vec and Glove only allow a single context-independent representation for each word.
To alleviate this problem, ELMo use vectors derived from a bidirectional LSTM \cite{hochreiter1997long} to obtain deep contextual word representations.
Different with previous methods, most recent pre-trained language models (e.g., BERT \cite{bert}, RoBERTa \cite{liu2019roberta}) use subword-based vocabularies and transformer \cite{vaswani2017attention} to learn context-dependent token representations.
Considering that word is basic lexical units and transformer is a powerful representation model, we are the first to explore combining their strengths to train a word-based BERT-style model using transformer.

Currently, sub-word approaches like WordPiece \cite{wu2016google} are widely used in pre-trained language models \cite{bert,liu2019roberta,lewis-etal-2020-bart}.
For convenience, most existing studies only adopt the widely-used vocabulary sizes, which typically range from 30K to 60K sub-word units, in implementation.
Recently, some researchers have begun to pay attention to the effects of vocabulary size.
Xu et al.~\shortcite{xu2021vocabulary} proposes VOLT, a vocabulary search approach without trial training, to find a well-performing small vocabulary in diverse settings.
Some researchers expand vocabulary size to 256K to train a GPT3-style \cite{brown2020language} model, named JURASSIC-1 \cite{lieber2021jurassic}.
The resulting vocabulary contains wordpieces, whole words, and multi-word expressions.
Our model differs from them in two ways.
First, our vocabulary only contains whole words without wordpieces or multi-word expressions.
Then we are the first to try to expand the vocabulary size to 1 million.

\section{Conclusion}
In this paper, we explore to train the BERT-style pretrained model over a word-based vocabulary without subword units using sampling-based training strategy.
We adopt various configurations, i.e., whether using pretrained word embeddings like Glove and different vocabulary size (e.g., 500K and 1M).
Results show that our model makes significant improvements on cloze test and machine reading comprehension.
What's more, WordBERT consistently achieves better $F_1$ scores on POS tagging, chunking and NER.
Model analysis illustrates that our models generate more diverse and meaningful words, while BERT tends to generate  general words.
Furthermore, we extend our approach to Chinese languages.
Results on several natural language understanding tasks indicate that our model significantly outperforms BERT. 
\bibliography{acl}

\begin{thebibliography}{26}
\expandafter\ifx\csname natexlab\endcsname\relax\def\natexlab#1{#1}\fi

\bibitem[{Anonymous(2021)}]{markbert}
Anonymous. 2021.
\newblock \href {https://openreview.net/pdf?id=7uE-SSLTgxw} {Markbert: Marking
  word boundaries improves chinese bert}.
\newblock \emph{ARR}.

\bibitem[{Bengio et~al.(2003)Bengio, Ducharme, Vincent, and
  Jauvin}]{bengio2003neural}
Yoshua Bengio, R{\'e}jean Ducharme, Pascal Vincent, and Christian Jauvin. 2003.
\newblock A neural probabilistic language model.
\newblock \emph{Journal of machine learning research}, 3(Feb):1137--1155.

\bibitem[{Bommasani et~al.(2021)Bommasani, Hudson, Adeli, Altman, Arora, von
  Arx, Bernstein, Bohg, Bosselut, Brunskill
  et~al.}]{bommasani2021opportunities}
Rishi Bommasani, Drew~A Hudson, Ehsan Adeli, Russ Altman, Simran Arora, Sydney
  von Arx, Michael~S Bernstein, Jeannette Bohg, Antoine Bosselut, Emma
  Brunskill, et~al. 2021.
\newblock On the opportunities and risks of foundation models.
\newblock \emph{arXiv preprint arXiv:2108.07258}.

\bibitem[{Brown et~al.(2020)Brown, Mann, Ryder, Subbiah, Kaplan, Dhariwal,
  Neelakantan, Shyam, Sastry, Askell et~al.}]{brown2020language}
Tom~B Brown, Benjamin Mann, Nick Ryder, Melanie Subbiah, Jared Kaplan, Prafulla
  Dhariwal, Arvind Neelakantan, Pranav Shyam, Girish Sastry, Amanda Askell,
  et~al. 2020.
\newblock Language models are few-shot learners.
\newblock \emph{arXiv preprint arXiv:2005.14165}.

\bibitem[{Cui et~al.(2019)Cui, Che, Liu, Qin, Yang, Wang, and Hu}]{cui2019pre}
Yiming Cui, Wanxiang Che, Ting Liu, Bing Qin, Ziqing Yang, Shijin Wang, and
  Guoping Hu. 2019.
\newblock Pre-training with whole word masking for chinese bert.
\newblock \emph{arXiv preprint arXiv:1906.08101}.

\bibitem[{Devlin et~al.(2018)Devlin, Chang, Lee, and Toutanova}]{bert}
Jacob Devlin, Ming{-}Wei Chang, Kenton Lee, and Kristina Toutanova. 2018.
\newblock \href {http://arxiv.org/abs/1810.04805} {{BERT:} pre-training of deep
  bidirectional transformers for language understanding}.
\newblock \emph{CoRR}, abs/1810.04805.

\bibitem[{Hochreiter and Schmidhuber(1997)}]{hochreiter1997long}
Sepp Hochreiter and J{\"u}rgen Schmidhuber. 1997.
\newblock Long short-term memory.
\newblock \emph{Neural computation}, 9(8):1735--1780.

\bibitem[{Lewis et~al.(2020)Lewis, Liu, Goyal, Ghazvininejad, Mohamed, Levy,
  Stoyanov, and Zettlemoyer}]{lewis-etal-2020-bart}
Mike Lewis, Yinhan Liu, Naman Goyal, Marjan Ghazvininejad, Abdelrahman Mohamed,
  Omer Levy, Veselin Stoyanov, and Luke Zettlemoyer. 2020.
\newblock \href {https://doi.org/10.18653/v1/2020.acl-main.703} {{BART}:
  Denoising sequence-to-sequence pre-training for natural language generation,
  translation, and comprehension}.
\newblock In \emph{Proceedings of the 58th Annual Meeting of the Association
  for Computational Linguistics}, pages 7871--7880, Online. Association for
  Computational Linguistics.

\bibitem[{Lieber et~al.(2021)Lieber, Sharir, Lenz, and
  Shoham}]{lieber2021jurassic}
Opher Lieber, Or~Sharir, Barak Lenz, and Yoav Shoham. 2021.
\newblock Jurassic-1: Technical details and evaluation.
\newblock \emph{White Paper. AI21 Labs}.

\bibitem[{Liu et~al.(2019)Liu, Ott, Goyal, Du, Joshi, Chen, Levy, Lewis,
  Zettlemoyer, and Stoyanov}]{liu2019roberta}
Yinhan Liu, Myle Ott, Naman Goyal, Jingfei Du, Mandar Joshi, Danqi Chen, Omer
  Levy, Mike Lewis, Luke Zettlemoyer, and Veselin Stoyanov. 2019.
\newblock Roberta: A robustly optimized bert pretraining approach.
\newblock \emph{arXiv preprint arXiv:1907.11692}.

\bibitem[{Mikolov et~al.(2013)Mikolov, Chen, Corrado, and
  Dean}]{mikolov2013efficient}
Tomas Mikolov, Kai Chen, Greg Corrado, and Jeffrey Dean. 2013.
\newblock Efficient estimation of word representations in vector space.
\newblock \emph{arXiv preprint arXiv:1301.3781}.

\bibitem[{Pennington et~al.(2014)Pennington, Socher, and
  Manning}]{pennington2014glove}
Jeffrey Pennington, Richard Socher, and Christopher Manning. 2014.
\newblock Glove: Global vectors for word representation.
\newblock In \emph{Proceedings of the conference on empirical methods in
  natural language processing}, pages 1532--1543.

\bibitem[{Peters et~al.(2018)Peters, Neumann, Iyyer, Gardner, Clark, Lee, and
  Zettlemoyer}]{peters-etal-2018-deep}
Matthew~E. Peters, Mark Neumann, Mohit Iyyer, Matt Gardner, Christopher Clark,
  Kenton Lee, and Luke Zettlemoyer. 2018.
\newblock \href {https://doi.org/10.18653/v1/N18-1202} {Deep contextualized
  word representations}.
\newblock In \emph{Proceedings of the 2018 Conference of the North {A}merican
  Chapter of the Association for Computational Linguistics: Human Language
  Technologies, Volume 1 (Long Papers)}, pages 2227--2237, New Orleans,
  Louisiana. Association for Computational Linguistics.

\bibitem[{Rajpurkar et~al.(2018)Rajpurkar, Jia, and
  Liang}]{rajpurkar-etal-2018-know}
Pranav Rajpurkar, Robin Jia, and Percy Liang. 2018.
\newblock \href {https://doi.org/10.18653/v1/P18-2124} {Know what you don{'}t
  know: Unanswerable questions for {SQ}u{AD}}.
\newblock In \emph{Proceedings of the 56th Annual Meeting of the Association
  for Computational Linguistics (Volume 2: Short Papers)}, pages 784--789,
  Melbourne, Australia. Association for Computational Linguistics.

\bibitem[{Rajpurkar et~al.(2016)Rajpurkar, Zhang, Lopyrev, and
  Liang}]{rajpurkar2016squad}
Pranav Rajpurkar, Jian Zhang, Konstantin Lopyrev, and Percy Liang. 2016.
\newblock Squad: 100,000+ questions for machine comprehension of text.
\newblock \emph{arXiv preprint arXiv:1606.05250}.

\bibitem[{Sang and De~Meulder(2003)}]{sang2003introduction}
Erik~F Sang and Fien De~Meulder. 2003.
\newblock Introduction to the conll-2003 shared task: Language-independent
  named entity recognition.
\newblock \emph{arXiv preprint cs/0306050}.

\bibitem[{Su(2020)}]{zhuiyiwobert}
Jianlin Su. 2020.
\newblock \href {https://github.com/ZhuiyiTechnology/WoBERT} {Wobert:
  Word-based chinese bert model - zhuiyiai}.
\newblock Technical report.

\bibitem[{Taylor(1953)}]{taylor1953cloze}
Wilson~L Taylor. 1953.
\newblock “cloze procedure”: A new tool for measuring readability.
\newblock \emph{Journalism quarterly}, 30(4):415--433.

\bibitem[{Vaswani et~al.(2017)Vaswani, Shazeer, Parmar, Uszkoreit, Jones,
  Gomez, Kaiser, and Polosukhin}]{vaswani2017attention}
Ashish Vaswani, Noam Shazeer, Niki Parmar, Jakob Uszkoreit, Llion Jones,
  Aidan~N Gomez, {\L}ukasz Kaiser, and Illia Polosukhin. 2017.
\newblock Attention is all you need.
\newblock In \emph{Advances in neural information processing systems}, pages
  5998--6008.

\bibitem[{Wang et~al.(2019)Wang, Shang, Liu, Lu, Liu, and Han}]{wang2019cross}
Zihan Wang, Jingbo Shang, Liyuan Liu, Lihao Lu, Jiacheng Liu, and Jiawei Han.
  2019.
\newblock Crossweigh: Training named entity tagger from imperfect annotations.
\newblock \emph{arXiv preprint arXiv:1909.01441}.

\bibitem[{Wu et~al.(2016)Wu, Schuster, Chen, Le, Norouzi, Macherey, Krikun,
  Cao, Gao, Macherey et~al.}]{wu2016google}
Yonghui Wu, Mike Schuster, Zhifeng Chen, Quoc~V Le, Mohammad Norouzi, Wolfgang
  Macherey, Maxim Krikun, Yuan Cao, Qin Gao, Klaus Macherey, et~al. 2016.
\newblock Google's neural machine translation system: Bridging the gap between
  human and machine translation.
\newblock \emph{arXiv preprint arXiv:1609.08144}.

\bibitem[{Xie et~al.(2018)Xie, Lai, Dai, and Hovy}]{xie-etal-2018-large}
Qizhe Xie, Guokun Lai, Zihang Dai, and Eduard Hovy. 2018.
\newblock \href {https://doi.org/10.18653/v1/D18-1257} {Large-scale cloze test
  dataset created by teachers}.
\newblock In \emph{Proceedings of the 2018 Conference on Empirical Methods in
  Natural Language Processing}, pages 2344--2356, Brussels, Belgium.
  Association for Computational Linguistics.

\bibitem[{Xu et~al.(2021)Xu, Zhou, Gan, Zheng, and Li}]{xu2021vocabulary}
Jingjing Xu, Hao Zhou, Chun Gan, Zaixiang Zheng, and Lei Li. 2021.
\newblock Vocabulary learning via optimal transport for neural machine
  translation.
\newblock In \emph{Proceedings of the 59th Annual Meeting of the Association
  for Computational Linguistics and the 11th International Joint Conference on
  Natural Language Processing (Volume 1: Long Papers)}, pages 7361--7373.

\bibitem[{Xu et~al.(2020)Xu, Hu, Zhang, Li, Cao, Li, Xu, Sun, Yu, Yu
  et~al.}]{xu2020clue}
Liang Xu, Hai Hu, Xuanwei Zhang, Lu~Li, Chenjie Cao, Yudong Li, Yechen Xu, Kai
  Sun, Dian Yu, Cong Yu, et~al. 2020.
\newblock Clue: A chinese language understanding evaluation benchmark.
\newblock \emph{arXiv preprint arXiv:2004.05986}.

\bibitem[{Yuan et~al.(2021)Yuan, Zhao, Du, Ding, Liu, Cen, Zou, Yang, and
  Tang}]{YUAN202165}
Sha Yuan, Hanyu Zhao, Zhengxiao Du, Ming Ding, Xiao Liu, Yukuo Cen, Xu~Zou,
  Zhilin Yang, and Jie Tang. 2021.
\newblock \href {https://doi.org/https://doi.org/10.1016/j.aiopen.2021.06.001}
  {Wudaocorpora: A super large-scale chinese corpora for pre-training language
  models}.
\newblock \emph{AI Open}, 2:65--68.

\bibitem[{Zhu et~al.(2015)Zhu, Kiros, Zemel, Salakhutdinov, Urtasun, Torralba,
  and Fidler}]{zhu2015aligning}
Yukun Zhu, Ryan Kiros, Rich Zemel, Ruslan Salakhutdinov, Raquel Urtasun,
  Antonio Torralba, and Sanja Fidler. 2015.
\newblock Aligning books and movies: Towards story-like visual explanations by
  watching movies and reading books.
\newblock In \emph{Proceedings of the IEEE international conference on computer
  vision}, pages 19--27.

\end{thebibliography}
\bibliographystyle{acl_natbib}
\appendix
\begin{table*}[ht]
\centering
\begin{tabular}{l|ccc|ccc|ccc}
\toprule
\multirow{2}{*}{\bf Dataset} & \multicolumn{3}{c|}{CLOTH-M} & \multicolumn{3}{c|}{CLOTH-H} & \multicolumn{3}{c}{CLOTH (Total)} \\ 

                 & Train & Dev  & Test & Train & Dev  & Test & Train & Dev   & Test      \\ \midrule 
                 \# passages            & 2,341  & 355  & 335  & 3,172  & 450  & 478  & 5,513  & 805   & 813    \\ 

\# questions           & 22,056 & 3,273 & 3,198 & 54,794 & 7,794 & 8,318 & 76,850 & 11,067 & 11,516  \\ 

\midrule

Avg. \# sentence            & & {16.26} & & & 18.92 & & & 17.79 & \\ 
Avg. \# words               &  & 242.88 &  & & 365.1 & & & 313.16 & \\ 
\bottomrule
\end{tabular}
\caption{The statistics of the training, development and test sets of CLOTH-M, CLOTH-H and CLOTH}
\label{tab:cloze_size} 
\end{table*}

\section{Pretraining}
\label{appendix-pretraining}
We use the following set of hyper-parameters to train models: total batchsize is 8,192 on 64x Tesla V100 GPUs and the max length is 512. We use Adam to update the parameters and set the learning rate to 5e-5 with a linear warmup scheduler.
We run the warmup process for 5k steps and train 200k steps in total.

\section{Fine-tuning}
\label{appendix-finetuning}
\subsection{Cloze Test}
Data statistics of the training/validation/testing data splits for CLOTH dataset are given in Table~\ref{tab:cloze_size}.
Following original experimental settings, we fine-tune for 2 epochs and a batch size of 4 and the learning rate of 1e-5.

\subsection{SQuAD}
For both SQuAD v1.1 and v2.0, we fine-tune for 2 epochs and a batch size of 48.
We select the best learning rate among \{ 1e-5, 3e-5 and 5e-5 \} on the development set.
Following the default settting, we set the max length as 384.

\subsection{POS Tagging, Chunking and NER}
We use the same hyper-parameters for three tasks.
We fine-tune for 10 epochs and set label smoothing factor as 0.1, max length as 200, batch size as 32, learning rate among \{ 1e-5, 3e-5 and 5e-5 \}.
We tune hyper-parameters and perform early stopping on the development set.
\subsection{Probing}
Data statistics of the number of masked words for different frequencies datasets are given in Table~\ref{tab:probing-data}.

\begin{table*}[]\setlength{\tabcolsep}{7pt}
    \centering   
    \begin{tabular}{l|c|c|c|c}
    \toprule
    \bf Datasets  & High & Medium & Low & Rare \\
        \midrule
    SQuAD  &  225,817 &	38,260 &	16,547 & 	4,002\\
WikiText-103 &1,336,055 &	229,805 &	99,328 & 8,567  \\
        \bottomrule
    \end{tabular}
    \caption{The number of masked words for different frequencies datasets.}
    \label{tab:probing-data}
\end{table*}

\end{document}